\documentclass{sigchi}

\makeatletter
\def\@copyrightspace{\relax}
\makeatother

\usepackage{balance}  
\usepackage{graphics} 
\usepackage{times}    
\usepackage{url}      
\usepackage{booktabs}
\usepackage[super]{nth}
\usepackage{float}
\usepackage{array}
\usepackage{booktabs}
\usepackage{amsmath}
\usepackage{tabularx}
\usepackage{epstopdf}
\makeatletter
\def\url@leostyle{%
  \@ifundefined{selectfont}{\def\UrlFont{\sf}}{\def\UrlFont{\small\bf\ttfamily}}}
\makeatother
\def\pprw{8.5in}
\def\pprh{11in}

\begin{document}

\title{Predicting Online Video Engagement Using Clickstreams}

\numberofauthors{3}
\author{
  \alignauthor Everaldo Aguiar\\
      \affaddr{University of Notre Dame}\\
   \affaddr{Notre Dame, Indiana 46556}\\
    \email{eaguiar@nd.edu}\\
  \alignauthor Saurabh Nagrecha\\
    \affaddr{University of Notre Dame}\\
    \affaddr{Notre Dame, Indiana 46556}\\
    \email{snagrech@nd.edu}\\
  \alignauthor Nitesh V. Chawla\thanks{Corresponding Author}\\
    \affaddr{University of Notre Dame}\\
    \affaddr{Notre Dame, Indiana 46556}\\
    \email{nchawla@nd.edu}\\
}

\maketitle

\begin{abstract}
In the nascent days of e-content delivery, having a superior product was enough to give companies an edge against the competition. With today's fiercely competitive market, one needs to be multiple steps ahead, especially when it comes to understanding consumers. Focusing on a large set of web portals owned and managed by a private communications company, we propose methods by which these sites' clickstream data can be used to provide a deep understanding of their visitors, as well as their interests and preferences. We further expand the use of this data to show that it can be effectively used to predict user engagement to video streams.  
\end{abstract}


\category{I.5.2.}{Pattern Recognition}{Design Methodology}

\section{Introduction}
The constant growth in volume, speed, availability, and functionality of the Web brings with it not only a variety of challenges and risks, but also a number of opportunities. While there have been a series of major advances in the field over time, one that has been given a considerable amount of attention in more recent years is that of \textit{personalization}. 

Data about users' online activity is continuously captured and analyzed. Advanced recommendation systems are now able to tell us what products we might be interested in buying, the books we will enjoy reading, what movies we should watch next, and even which diseases we are at risk of contracting. From a business perspective, the benefits of being able to understand customers in this level of detail are unquestionable.

Methods for capturing user data on the Web are also becoming increasingly efficient. As described in \cite{mob00}, the browsing behavior of individual users can be recorded at the granularity of mouse clicks with little to no work needing to be done. A number of services, both free and proprietary, offer user tracking solutions that can be implemented and deployed within minutes. However, the feedback that one usually gets from these tools is often in the form of simplistic aggregate statistics that do not offer a deeper understanding of user behavior. 

With that in mind, we set to analyze the application of some of these ideas to a specific context, while having as our major goal the understanding of each user as an individual unit. For this study, we were provided a large dataset that describes user clicks generated within a two-month span and across a number of websites managed by a large communications company.

This paper describes the process through which we parsed, analyzed, and drew knowledge from a user-generated clickstream dataset provided by a large communications company. We begin by showing, from a more general perspective, how this type of data can be used to identify particularly interesting trends in user interest, and to further illustrate the usefulness of this information, we describe how we applied methods to predict user engagement to video streams and discuss their accuracy.

\begin{figure}[h]
  \centering
    \includegraphics[width=0.5\textwidth]{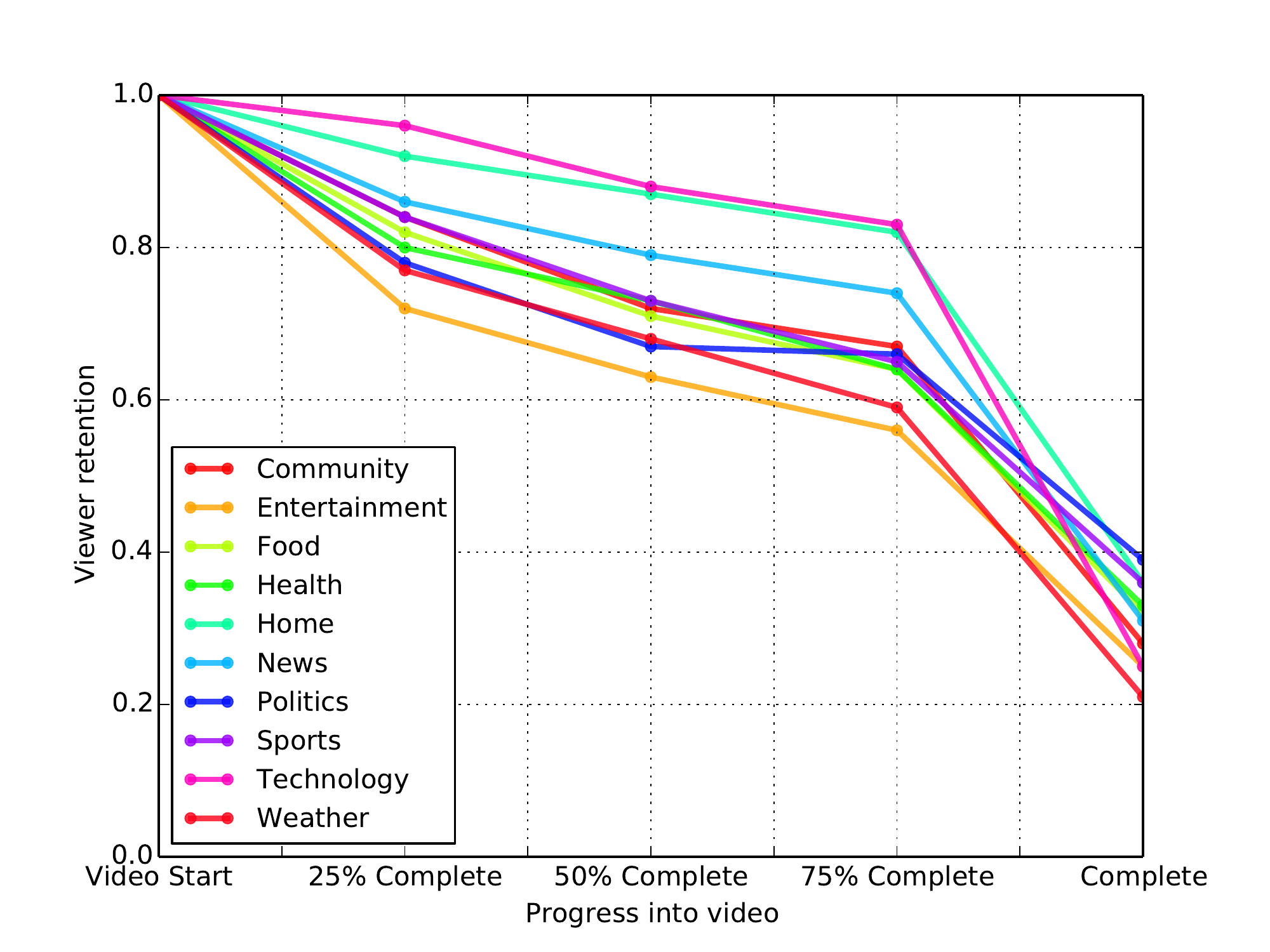}
    \caption{Video viewership drop-off by category of content}
    \label{fig:drop-off}
\end{figure}

Distributing content that entices user engagement and captures large audiences is the ultimate goal of all web media providers. Measuring and forecasting these variables, however, is not an easy task. As Figure \ref{fig:drop-off} illustrates, as time goes by, the amount of users that remain tuned to video streams dramatically decreases. For certain categories, the percentage of users that actually watch videos to completion can be as low as 20\%.

To address this undesired outcome, we propose the development of clickstream-based models that can learn the individual preferences and characteristics of each user, and utilize this information to predict how ``engaged" they will be to a particular video stream. Being able to know, in advance, if a user is likely to exit a video prematurely allows content providers some leeway to implement personalized intervention strategies aimed at maximizing viewership retention.

The remaining portion of the paper is organized as follows -- The next section gives an overview of the most recent related literature. That is followed by a detailed coverage of clickstream data representation and a description of our particular dataset. We then elaborate on the methods applied in this study, the results obtained, and their importance. Finally, the last section draws conclusions about this exercise and argues for the latent potential that resides in user-generated clickstream data.

\section{Related Work}
Interest in analyzing the online activities of users is as old as providing consumable content itself. This problem has piqued the interest of multiple fields, namely marketing, psychology and computer science. 


Since user activity provides an immense amount of measurable secondary data, various models to predict multiple aspects of their behavior have been proposed. User interaction has been studied at various levels--- from gaze tracking~\cite{dreze2003internet} to broader patterns of path traversal within a website~\cite{banerjee2001clickstream,montgomery2004modeling}. Simple duration and dwell-time~\cite{buk03} can be used to predict when a user exits the site. User classification ~\cite{moe2003buying} can be used to identify what the user is specifically looking for and even morph the website~\cite{hauser2009website} according to the custom tastes of that particular user profile. Personalized content based on click history has been implemented and widely adopted by commercial content providers~\cite{das2007google,liu10}.

With the distribution of video content online becoming mainstream, the way we study user engagement has been greatly enriched. Studies like~\cite{dob11} have measured the role of video content quality in influencing user engagement, but did not utilize clickstreams to contextualize the video views. Online video engagement for Massive Open Online Courses (MOOCs)~\cite{guounderstanding} has shown that the lessons learned from analyzing video views can be used to improve video authoring, editing and interface design. It also emphasizes the value of \emph{video dropout} as a metric for engagement. Though the MOOC work lacks the contextual history of the users, in this paper we leverage similar and many other clickstream features to predict video engagement.

\begin{figure}[h]
  \centering
    \includegraphics[width=0.4\textwidth]{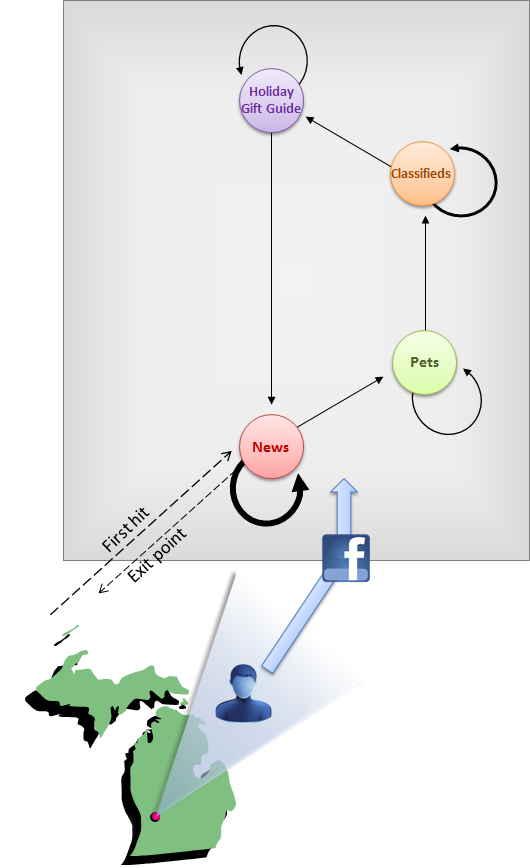}
    \caption{A simple illustration of the clickstream of a typical user}
    \label{fig:cstream}
\end{figure}

\section{Clickstream Data Representation}

Clickstream data consists of a ``virtual trail" that users leave behind while they interact with a given system, website or application. More specifically, data that describes the state of a user's current session is recorded each time a click is performed, and the aggregation of that produces a \textit{clickstream}, which can be used to reconstruct all actions taken by the user while he or she utilized that given product.

While applicable to a variety of scenarios, the collection and analysis of clickstreams has become most notably popular in the context of Web-based tools and websites. As highlighted by Srivastava et. al.~\cite{sri00},  the analysis of such information has potential applications in a number of areas such as website personalization and modification, system improvement, business intelligence and usage characterization. Our contributions fall mainly within the first and last domains.

\begin{table*}[t]
\centering
\begin{tabular}{cll}
\hline
Feature Type & Feature Name        & Description                                                      \\ \hline
\textit{Nominal}      &                     & \\
             & Browser             & The browser that was used                                        \\
             & Channel             & The site that the page view belongs to                           \\
             & City                & The city the user accessed the page from                         \\
             & Cookies             & Whether the user had cookies turned on or not                    \\
             & Country             & The country the user accessed the page from                      \\
             & Domain              & Domain of the user's ISP                                         \\
             & Exclude hit         & Identifies web crawlers                                          \\
             & First hit page      & URL the user first landed on the website                         \\
             & Frequency of visits & Denotes hourly, daily, weekly, monthly or yearly visit           \\
             & IP                  & Refers to the IP address of the user                             \\
             & New visit           & Determines whether the user is new to the site, based on cookies \\
             & Referrer            & Lists the URL of the website that referred this user             \\
             & Region              & Refers to the state or region the user was in                    \\
             & Search Keywords     & The search string which led to the particular page               \\
             & Section             & The section of the website where the click took place            \\
             & Subsection          & Subsection of the website where the click took place             \\
\textit{Numeric}      &                     & \\
             & First hit time      & Timestamp of when the user first landed on the website           \\
             & Last click          & Time stamp of when the last click was  made by the user          \\
             & Last visit          & Refers to when the user visited the site last                    \\
             & Time \& Date        & Timestamp of when the click instance happened                    \\
             & Visit number        & Refers to the number of times the user has visited the site      \\ \hline
\end{tabular}
\caption{Dataset features described}
\label{tb:features_description}
\end{table*}


\subsection{Our Dataset}

The data we utilized for this study was provided to us by a large U.S.-based communications company that operates in the radio, TV, newspaper and online media domain. They manage a few dozens of websites, all of which are embedded with clickstream capturing functionality. Next, we give a detailed description of the most important features this dataset contains.

User activity is continuously captured by numerous servers across the country and is then concatenated at the end of the day in the form of daily ``dumps". We utilized 59 of these files that covered the period ranging from December 4, 2012 to January 31, 2013. Altogether, these files contain an upwards of 65 million click instances.

Each click instance recorded is characterized by a large number of \textit{features} (161 in this case). Table \ref{tb:features_description} lists a small subset of the most relevant features and a brief description of each. With that information we are able to determine (1) how users reached the website, (2) what attracted them there, (3) what actions they performed while on the site and (4) how they eventually exited.

Note that while there is no feature that captures the event of a user leaving the website, as is common practice, we work under the assumption that when a user is inactive for a period longer than 30 minutes (i.e., no click events originate from this person during that time), we simply say that the user has exited the site.

This assumption allows us to group these click events from the original datasets into user sessions, which illustrate the path a user takes while browsing the website and can be used to identify areas that attract more (or less) traffic. 

Figure \ref{fig:cstream} illustrates one individual session chosen at random from our dataset. We can see that the user in this case was referred to our domain through a link that he or she found on a social network website and that their visit consisted of several hops, most of which happened in the \textit{news} section.

\begin{figure}[h]
  \centering
    \includegraphics[width=0.5\textwidth]{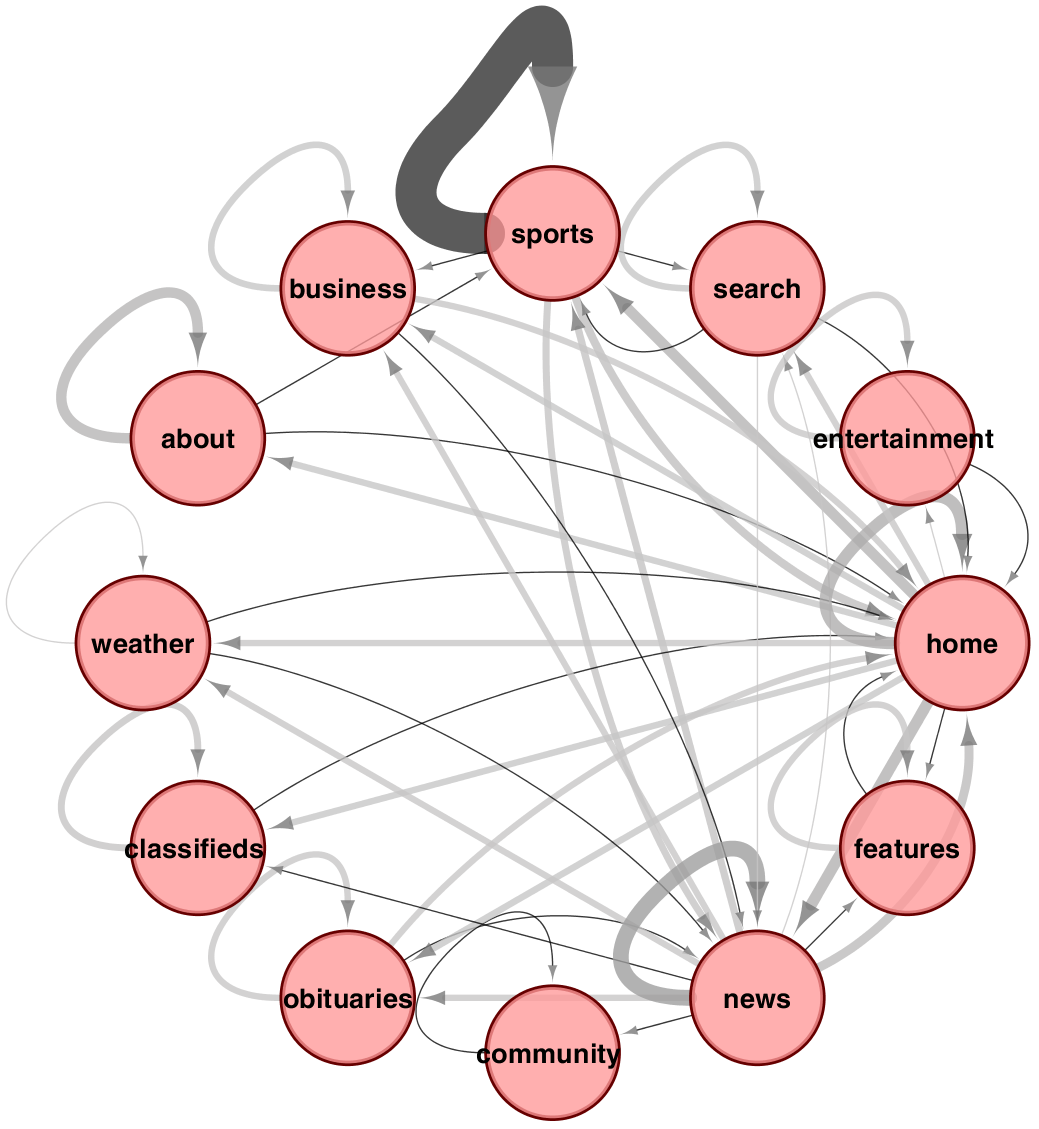}
    \caption{Clickstream network for a news-media website. The various nodes displayed here represent different sections. The direction of the arrows represents user traffic flowing between these sections and the thickness is indicative of volume of said traffic.}
    \label{fig:siteactivity}
\end{figure}

Aggregating these sessions allows us to visualize which areas of the website are more popular, as well as which links connecting different sections are traversed the most. Take for instance the example illustrated in Figure \ref{fig:siteactivity}. To generate this particular graph, we isolated the sessions corresponding to a certain newspaper's website, its 12 most popular sections, and the traffic between them. Among other observations, we noticed that the readers of this particular newspaper were often prone to navigating to the \textit{sports} section and reading multiple articles there.

Furthermore, these sessions can be aggregated, producing a high-level view of the entire website structure by popularity of section. Figure~\ref{fig:siteactivity} illustrates this concept.

Lastly, we note that based on information retrieved from specific features of our dataset, it is possible to determine if a user is simply browsing text articles, displaying image galleries or streaming online video. The following sections of this paper will describe how we used this fact to aid in the development of predictive models for video viewership engagement.


\section{Methods}
\subsection{Identification of Video Exit Instances}
When a user watches a video, a separate log entry is made corresponding to when he or she completes watching a certain percentage of the video, while a player ID remains constant. This makes the clickstream log reflect a cumulative history of the viewer's progress within that video.

By filtering the data to get only clicks corresponding to video instances, and then by IP address, we obtain the entire video viewing activity of each IP. From this modified dataset, we isolate an individual ``video view" table by specifying the player ID. This table is then sorted chronologically and filtered by session timeout. The last entry corresponds to the viewer's exit point. This gives us a unique session for a visit. In combination with the current session data, and data from cookies, we retrieve the user's unique historical browsing patterns. It should be noted that in the absence of cookies, we treat the user as a fresh incoming visitor. For our analysis, we isolated only the instances where the user exited the video.
Due to the inherently discretized nature of the data collection, we get a coarse-grained estimate of when the user reached a certain percent of the video. If the last entry shows that the user watched 50\% of a video, it can be inferred that the user exited at \(p\%\), such that \(p\in [50,75)\).

\subsection{Feature Selection}

Using various feature selection methods, we reduced the size of our dataset from the original 161 features to the 12 best descriptors. Among these were features like IP, location, content annotations, and referrer information. 
Out of the 161 features in a typical video exit instance, 40 are mutually redundant, and 32 are constant in value. This motivates the need to find a set of features that is the best descriptor of the target class (in this case, the percent of video the user watches before exiting)~\cite{guyon2003introduction}. 
We investigated various feature selection methods which support mixed data types and ranked the top features. One would expect these features to encompass measurable user traits which influence their interest in the video.

Various feature selection methods aim to remove redundant and irrelevant features using different statistical means, which have their respective strengths. Though a popular choice in machine learning, correlation based feature selection (CFS) was not considered due to the sparse nature of the data~\cite{hall1999correlation}. A more detailed study of these can be found in~\cite{yang1997comparative,forman2003extensive}. The feature selection methods employed in this problem are described below:
 
\subsubsection{Chi Squared} 
The chi squared (\(\chi^2\)) method measures how much deviation is seen in the observed data from the case where the class value and the feature are independent of each other. It evaluates whether the feature and class occurrences are randomly related, or exhibit some relation.
\subsubsection{Information Gain} 
Information gain~\cite{quinlan1986induction} measures how much entropy is lost when the feature is present vs. absent. 
\subsubsection{Gain Ratio}
Information gain favors attributes with many values over those with fewer values, the gain ratio~\cite{quinlan1993c4} compensates for this by factoring in the amount of split caused by the feature.
\subsubsection{One R}
One R formulates a set of simple relationships between the features and ranks the features based on how accurate these rules are.
\subsubsection{Symmetric Uncertainty}
Symmetric uncertainty~\cite{witten2005data, eom2004pubminer} targets attributes which correlate well with the class but have little intercorrelation.

The results of these feature selection methods are summarized in Table~\ref{tab:FeatureSelection}. The attributes in the table are the ones which consistently appear in the top 10\%. These are the attributes which influence video exit points the most. 
\newcolumntype{M}[1]{>{\centering\arraybackslash}m{#1}}
\begin{center}
\begin{table}
\centering
    \begin{tabular}{lM{7mm}M{7mm}M{7mm}M{7mm}M{7mm}}
    \hline
    \textbf{Features}                      & \textbf{Chi} & \textbf{IG} & \textbf{GR} & \textbf{oneR} & \textbf{Symm} \\ \hline
    Time                   & 1          & 1        & 7         & -    & 2       \\
    IP                     & 2          & 2        & 9         & -    & 3       \\
    First hit referrer     & 3          & 3        & 5         & 2    & 5       \\
    First hit page         & 4          & 5        & 10        & -    & 7       \\
    Story title            & 5          & 4        & 2         & 1    & 1       \\
    Search engine         & 6          & 7        & 3         & 3    & 8       \\
    City                   & 7          & 6        & -         & -    & 9       \\
    ISP                    & 8          & 8        & -         & -    & 10      \\
    Referrer type          & 9          & 10       & 1         & -    & 4       \\
    \# Pages viewed        & 10         & 9        & 8         & -    & 6       \\
    Search page num        & -          & -        & 4         & 4    & -       \\
    Frequency of visits    & -          & -        & 6         & 5    & -       \\ \hline
    \end{tabular}
    \caption{\textbf{Feature Selection Rankings}. (description of abbreviations)}
    \label{tab:FeatureSelection}
\end{table}
\end{center}

The time of viewing influences at what point people are prone to exit the video. IP address, in conjunction with location, and ISP indicate who is watching the video and thus offer a personalized facet to the prediction. The number of pages viewed by a person and frequency of visits can be perceived to be reflective of the person's interest in the site. The referrer which brought the viewer on the site can influence the engagement of the viewer; a viewer coming from a social network link interacts differently than one who had the site bookmarked on their browser. The entry point is the first page the viewer saw in their current viewing session; this determines their interest in consuming further content. The actual title of the story includes the section which the video is under. As we had observed in Figure~\ref{fig:drop-off}, users viewing ``Technology" related videos were less likely to exit than those viewing ``Entertainment" related videos. 


\subsection{Classification}

Our aim is to predict how much of the video a user watches before exiting. In our dataset, we find that this is represented by 5 distinct markers, which correspond to the percentage of the video the user watched before exiting. We formulate two classification tasks- to predict what percent of the video is watched, and whether the user exits the video ``early" (before reaching 50\% of the video).

This prediction task involving 5 classes. Since it is relevant to predict users who exit early on in the video, we assume that users who exit the video at the beginning or having viewed 25\% of the video to have exited ``early".  As described above, this would correspond to users who have viewed 0 to 49\% of the video. 

\begin{figure}[bh]
  \centering
    \includegraphics[width=0.42\textwidth]{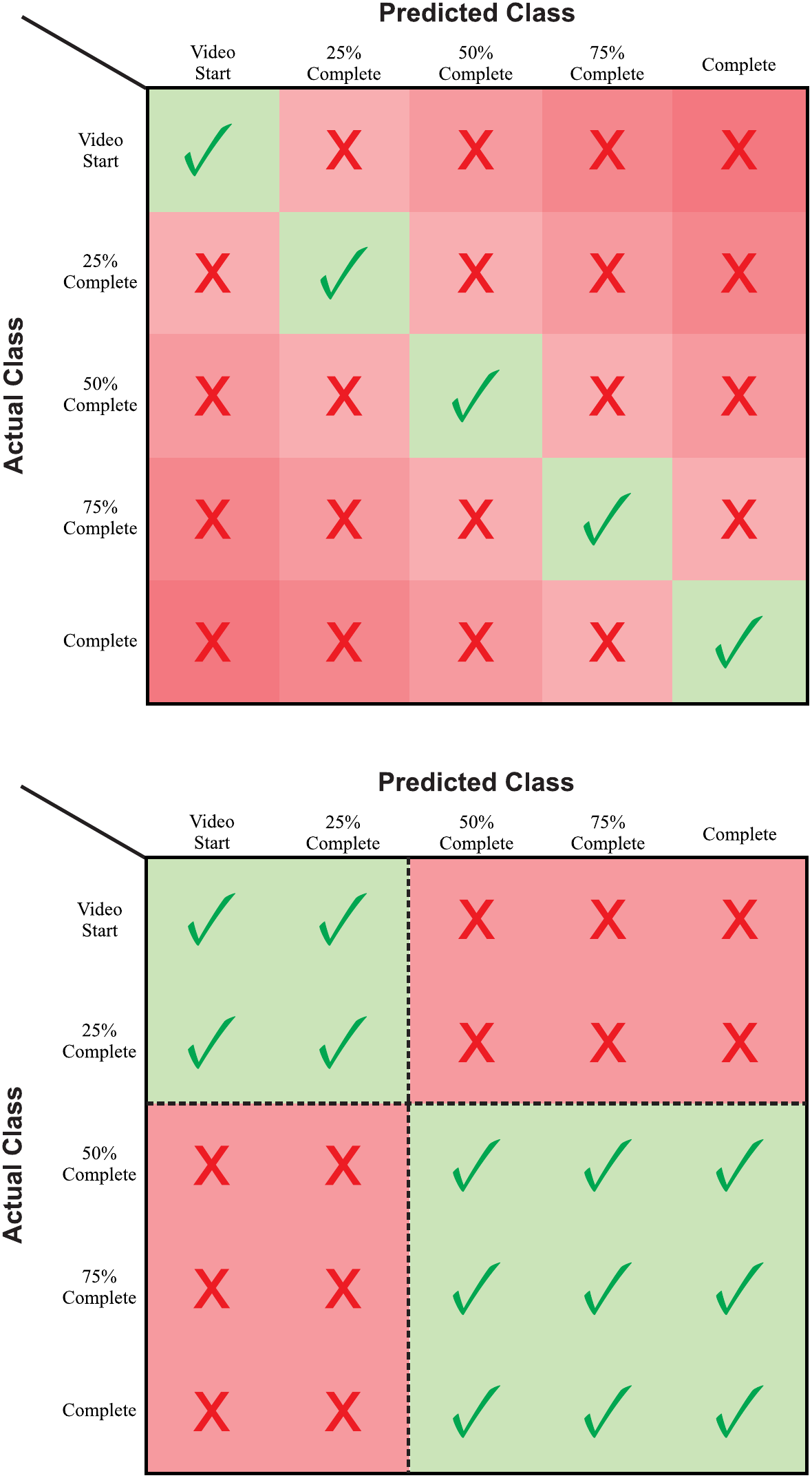}
    \caption{Converting the Percentages Classification to Early Exit Classification: The 5 class problem (top) is reduced to a binary classification problem by merging classes (bottom).}
    \label{fig:classmerger}
\end{figure}

We can refine the problem as the binary prediction of these ``early exits". The classes would then be a merger of the previously mentioned 5 classes, with the first two combined to form that of ``early exits" and the latter 3 being those who chose to not exit early. This simplification is depicted in the representative expected confusion matrices for both of the classification tasks as displayed in Figure \ref{fig:classmerger}. We have performed both prediction analyses on our data.

\subsubsection{Naive Bayes}

Among the simplest and most primitive classification algorithms, this probabilistic method is based on the Bayes Theorem~\cite{bay63} and strong underlying independence assumptions. That is, each feature is assumed to contribute independently to the class outcome. 

\subsubsection{C4.5 Decision trees}

C4.5 Decision Trees~\cite{quinlan1993c4} work by building a tree structure where split operations are performed on each node based on information gain values for each feature of the dataset and the respective class. At each level, the attribute with highest information gain is chosen as the basis for the split criterion.

\subsubsection{Repeated Incremental Pruning to Produce Error Reduction}

RIPPER~\cite{cohen1995fast} is a  rule based classification tree learner. It is algorithmically faster than C4.5, having a complexity of \(O(n(log(n))^2)\) as opposed to C4.5's complexity of the order \(O(n^3)\). RIPPER constructs an initial set of rules and then iteratively optimizes it according to a tunable parameter. It is implemented in Weka under the ``JRip class".

\subsubsection{Random forests}

Random forests~\cite{bre01} combine multiple tree predictors in an ensemble. New instances being classified are pushed down the trees, and each tree reports a classification. The ``forest" then decides which label to assign to this new instance based on the aggregate number of votes given by the set of trees.

\subsubsection{Decision Tables} 
 
Decision Table classifiers \cite{koh95} are built by concatenating a series of rules derived from the feature set to corresponding class outcomes. This method as its major advantages the fact that it is easy to interpret and notably efficient.

\subsubsection{Random Subspaces}

The random subspace method~\cite{ho1998random} is an ensemble classifier whose individual classifiers operate on random subsets of the feature set. The predictions made by the individual classifiers are combined using the posterior probabilities of each class in the constituent classifiers. This method looks at the classification problem from various perspectives by randomizing the selection of features. 

\subsubsection{Stacking}
Stacking~\cite{wol92} is a meta-classification scheme which employs an ensemble of classifiers and performs the learning task on two levels. First, the classifiers in the ensemble are trained on the data, then the meta-classifier learns from \emph{their} predictions and the training labels of the data.

\subsection{Key Performance Indices / Metrics Utilized}
Our key performance index is the accuracy of prediction of when the user will drop-off in the video. To obtain these predictions, we perform 10-fold cross-validation on the available data using various classification methods. In 10-fold cross validation, the data is randomly partitioned into 10 subsets and predictions are made on each of these. These predictions are then aggregated to provide the overall performance of the classifier,
which we measured by the accuracy and area under the Receiver Operating Characteristics curve (AUROC)
, all of which are described in further detail below. Each of these measures depict various aspects of the prediction results.

\subsubsection{Accuracy}
The accuracy of a classifier is perhaps the simplest measurement of its performance. It represents the percentage of total instances that were correctly classified. We would like for this to be as high as possible. The baseline for accuracy is that of a perfectly random prediction. For a binary classification problem, this would be 50\% and for a 5 class problem, the baseline accuracy would be 20\%. Any classifier which delivers statistically greater accuracy than these respective baselines, is considered to be better than a random predictor.

\subsubsection{Receiver Operating Characteristics (ROC) curves}
A system tuned to increase accuracy does not necessarily make it a good predictor. Relying on accuracy alone does not provide insights into the nature of misclassified instances. ROC curves~\cite{fawcett2006introduction} are a way to quickly compare multiple classifiers. The goal of a classifier in ROC space is to be as close to the upper-left corner as possible. In ROC space, if the curve for one classifier is closer to the upper-left corner than that for another, then it is considered to have a superior performance.





\begin{table}
  \centering
    \begin{tabular}{l M{13mm}M{13mm}M{13mm}} \toprule
        Dataset   & Classifier   & Acc & AUROC \\ \midrule
			& NB & 0.416 & \textbf{0.718} \\ \cmidrule{2-4}
			& C4.5 & 0.547 & 0.699 \\ \cmidrule{2-4}
		Multiclass	& RIP & 0.547 & 0.629 \\ \cmidrule{2-4}
			& DT & 0.543 & 0.717  \\ \cmidrule{2-4}
        		& ST & \textbf{0.569} & 0.652  \\ \midrule
        		
			& NB & 0.772 & 0.753 \\ \cmidrule{2-4}
			& C4.5 & 0.809 & 0.794  \\ \cmidrule{2-4}
		Binary	& RIP & 0.826 &  0.697  \\ \cmidrule{2-4}
			& DT & 0.806 &   \textbf{0.805}  \\ \cmidrule{2-4}
        		& ST & \textbf{0.846} &  0.739  \\ \bottomrule   
    \end{tabular}
    \caption{Summary of results obtained for each classifier and dataset. The classifiers used are NB: Naive Bayes, C4.5: C4.5 decision tree, RIPPER: Repeated Incremental Pruning to Produce Error Reduction, DT: Decision Table, ST: Stacking using random subspaces of decision trees}
    \label{tab:acc}
\end{table}

\section{Experimental Results}

We evaluated the performance of each of the classifiers used, with 10-fold cross validation for both the multiclass and the binary classification predictions. Table~\ref{tab:acc} summarizes the results of all experiments.

\subsection{Multiclass Prediction}
We see that in terms of sheer accuracy, the stacked classifiers performed slightly better than other methods, achieving an accuracy of 56.9\%. In terms of AUROC, however, it is seen that Naive Bayes performs much better, closely followed by Decision Tables. These simple classifiers might not have the best accuracy, but outperform the others.

\subsection{Binary Class Prediction}
In this second scenario, we associate a semantic meaning to the drop-off percentage point and predict if the user will exit early or not. This refinement of the problem statement gives us a much better performance across the board. The stacked classifiers, for instance, achieve a remarkable accuracy of 84.6\% when predicting which users exited their video streams prematurely. As it was the case with the multiclass problem, we again saw that Decision Tables and Naive Bayes surpassed the other classifiers in terms of AUROC values. 

Though stacked classifiers give greater accuracy, they are not as good as Decision Tables or Naive Bayes in predicting early drop-off. This is still reflective of the general trends observed in the multiclass problem as we have merely merged classes, the underlying data remains the same.

In both, the multiclass and binary class prediction, it is observed that simpler rule based learners outperform complicated meta-classifiers. This is documented in~\cite{dvzeroski2004combining}, showing that stacking does not always outperform the best classifier.

\begin{figure}[H]
  \centering
    \includegraphics[width=0.45\textwidth]{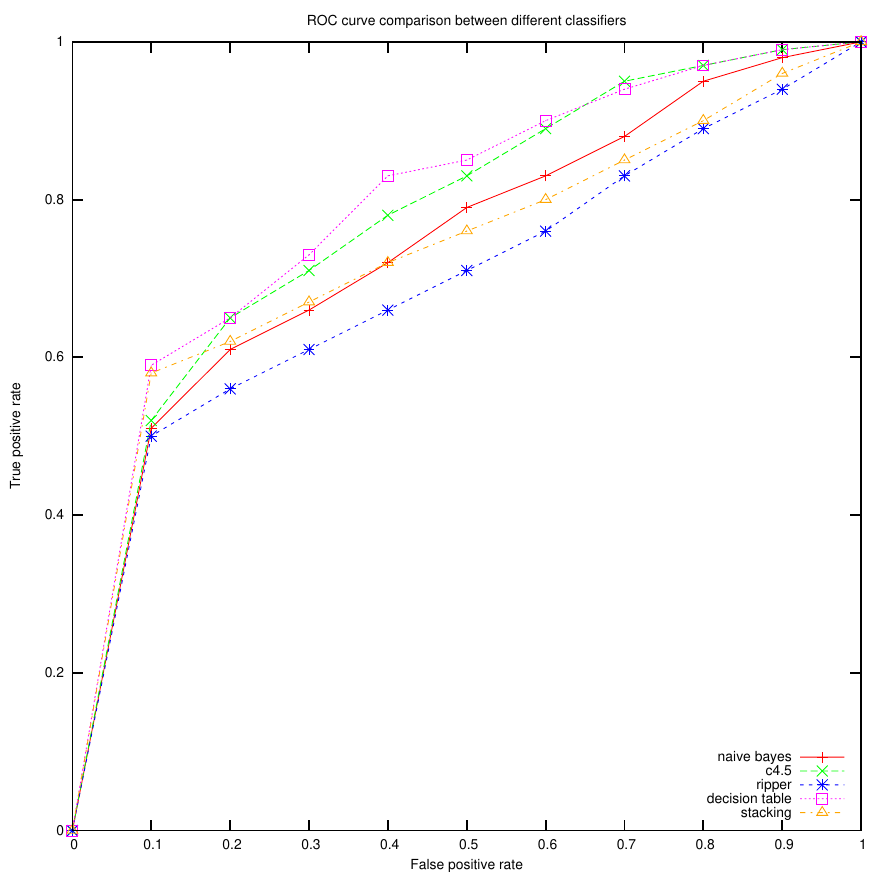}
    \caption{ROC curves for the binary class problem. A comparison of various classifiers to predict early exit behavior.}
    \label{fig:precRecall}
\end{figure}

We see that simple classification algorithms can be used to achieve comparable, or even better performance than complicated meta-classifiers. Besides the obvious performance superiority, it is desirable to use simpler classifiers on grounds of computational complexity, as implementing these is algorithmically more scalable and thus offers faster runtime.

\section{Conclusions}

We demonstrated how clickstream data can be used to predict ``early exits" in online videos. By constructing models to this effect, we were able to identify with high accuracy which video streaming sessions are likely to terminate prematurely. Additionally, we compared and contrasted the performance of a number of classifiers, highlighting those that we found to be particularly fit to this problem. Having knowledge of such information would allow content providers to personalize how their media is distributed so as to increase user retention, and as a result, business value.

\balance

\bibliographystyle{acm-sigchi}
\bibliography{sample}
\end{document}